\documentclass[lettersize,journal]{IEEEtran}
\usepackage{amsmath,amsfonts}
\usepackage{algorithmic}
\usepackage{algorithm}
\usepackage{array}
\usepackage[caption=false,font=normalsize,labelfont=sf,textfont=sf]{subfig}
\usepackage{textcomp}
\usepackage{stfloats}
\usepackage{url}
\usepackage{verbatim}
\usepackage{graphicx}
\usepackage{cite}
\usepackage{listings}
\usepackage{hyperref}
\usepackage{multicol}
\usepackage{multirow}
\usepackage{makecell}
\usepackage{booktabs}

\usepackage{pifont}
\usepackage{color}
\usepackage{xcolor}
\usepackage{colortbl}

\definecolor{lightgray}{rgb}{0.8, 0.8, 0.8}
\definecolor{lgray}{rgb}{0.66, 0.66, 0.66}

\definecolor{lblu_tab}{RGB}{225, 235, 246}
\definecolor{orange_vitad}{RGB}{222, 131, 68}
\definecolor{blue_vitad}{RGB}{106, 153, 208}
\definecolor{trajectory_green}{RGB}{126, 171, 85}
\definecolor{trajectory_yellow}{RGB}{245, 194, 66}
\definecolor{tab_others}{RGB}{235, 235, 235}
\definecolor{tab_ours}{RGB}{225, 235, 246}

\definecolor{whit_tab}{RGB}{255, 255, 255}
\definecolor{gray_tab}{RGB}{246, 246, 246}
\definecolor{oran_tab}{RGB}{252, 242, 237}
\definecolor{blue_tab}{RGB}{227, 240, 251}

\newcommand{\cmark}{\ding{52}}

\newcommand{\xmark}{\ding{56}}

\definecolor{linkcolor}{RGB}{255,0,0}
\definecolor{urlcolor}{RGB}{255,105,180}
\definecolor{citecolor}{RGB}{66,168,235}

\begin{document}

\title{Semantic Frame Interpolation}

\author{Yijia Hong, Jiangning Zhang, Ran Yi, Weijian Cao, Xiaobin Hu, Lizhuang Ma, Shuicheng Yan % <-this % stops a space
}
\markboth{Journal of \LaTeX\ Class Files,~Vol.~14, No.~8, August~2021}%
{Shell \MakeLowercase{\textit{et al.}}: A Sample Article Using IEEEtran.cls for IEEE Journals}


\maketitle

\begin{abstract}
Generating intermediate video content of varying lengths based on given first and last frames, along with text prompt information, offers significant research and application potential. However, traditional frame interpolation tasks primarily focus on scenarios with a small number of frames, no text control, and minimal differences between the first and last frames. Recent community developers have utilized large video models represented by Wan to endow frame-to-frame capabilities. However, these models can only generate a fixed number of frames and often fail to produce satisfactory results for certain frame lengths, while this setting lacks a clear official definition and a well-established benchmark. In this paper, we first propose a new practical \textbf{S}emantic \textbf{F}rame \textbf{I}nterpolation (SFI) task from the perspective of academic definition, which covers the above two settings and supports inference at multiple frame rates. To achieve this goal, we propose a novel SemFi model building upon Wan2.1, which incorporates a Mixture-of-LoRA module to ensure the generation of high-consistency content that aligns with control conditions across various frame length limitations. Furthermore, we propose SFI-300K, the first general-purpose dataset and benchmark specifically designed for SFI. To support this, we collect and process data from the perspective of SFI, carefully designing evaluation metrics and methods to evaluate the performance of the model in multiple dimensions, including images and videos, and various aspects, including consistency and diversity. Through extensive experiments on SFI-300K, we demonstrate that our method is particularly well-suited to meet the requirements of the SFI task. 
\end{abstract}


\section{Introduction} \label{sec:introduction}

\begin{figure*}
  \centering
  \includegraphics[width=\textwidth]{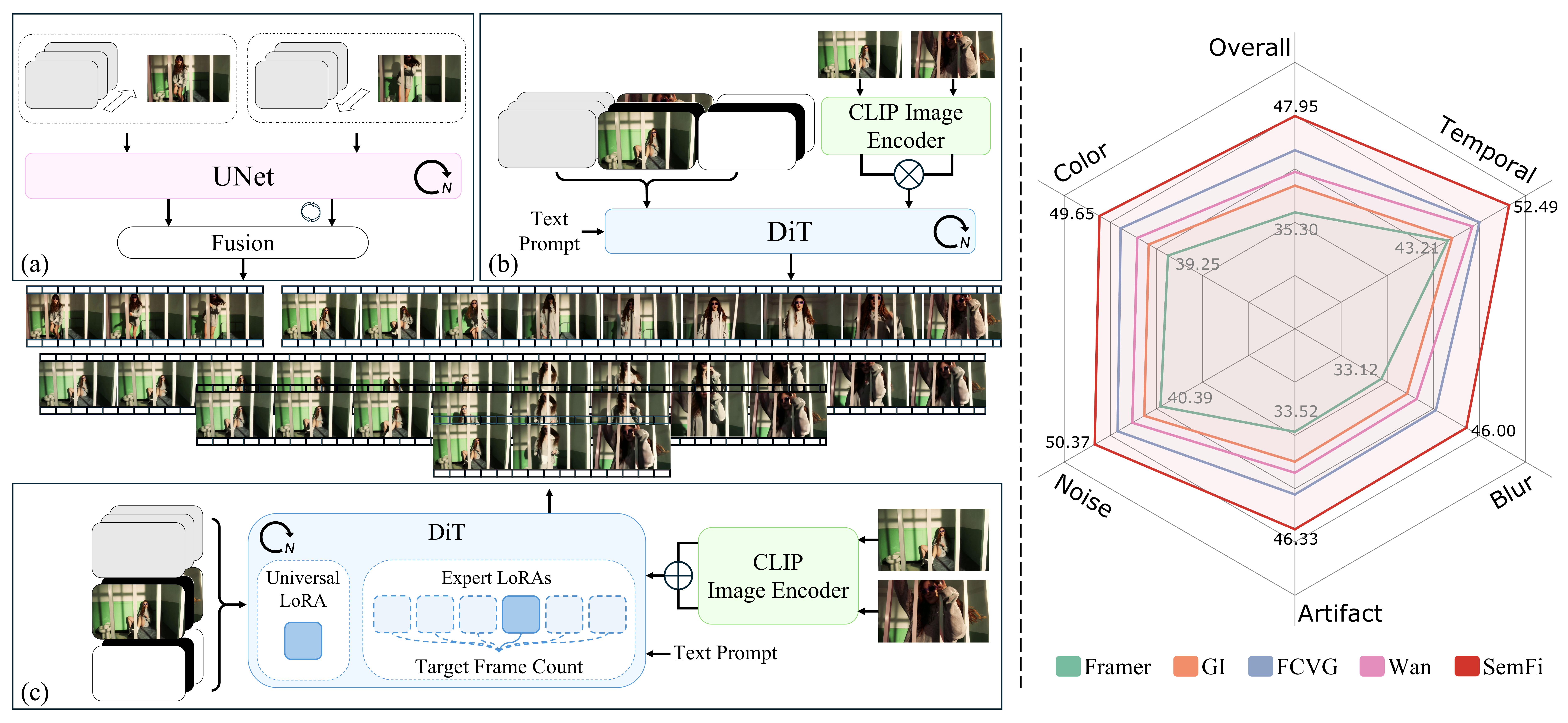}
  \caption{\textbf{Left:} (a) Conventional video frame interpolation approaches (illustrated here with an SVD-aided method) are limited to producing a small number of frames or low-density information; (b) Foundation-video-model-based frame-to-frame frameworks exhibit a stronger bias toward long-range generation, but are suboptimal for short-term frame generation. (c) Our SemFi approach utilizes a single, unified model to generate versatile intermediate frames that maintain semantic coherence with text descriptions. \textbf{Right:} Standardized evaluation results. Our method demonstrates superior performance on semantic frame interpolation tasks compared to prior video interpolation methods and frame-to-frame frameworks.}
  \label{fig:1}
\end{figure*}

\IEEEPARstart{S}{ynthesizing} intermediate frames between two given images has significant practical value in applications such as video editing, animation production, and augmented reality, where smooth transitions and semantically consistent content creation are crucial. Recent approaches for this task can be broadly categorized into two paradigms. Conventional video frame interpolation (VFI) methods~\cite{reda2022film, ding2024video,prasanna2024video,cheng2021multiple, briedis2023kernel,wang2025framer, wang2024generative, zhu2024generative, yang2025vibidsampler} are effective for short-range interpolation with small motion, but they are typically limited to a small number of intermediate frames and provide no semantic controllability. In contrast, recent foundation-video-model-based (FVM-based) frame-to-frame generation methods~\cite{kelin, wang2025wan} can handle larger temporal gaps and produce semantically richer transitions, yet they are often suboptimal for precise short-range interpolation, leading to artifacts or misalignments. These limitations lead to two key challenges: \textbf{\textit{1)}} \textit{achieving high visual fidelity while preserving precise instruction-following, so that the generated content remains consistent with both low-level motion and high-level user intent}; and \textbf{\textit{2)}} \textit{enabling robust multi-scale generation, i.e., building a unified framework that can adapt to both short-term precision and long-term coherence without sacrificing perceptual quality or temporal stability across different frame counts}.

To address these challenges, we present a comprehensive solution spanning task formulation, dataset construction, and methodological innovation. 
First, we propose \textbf{Semantic Frame Interpolation (SFI)}, a novel and generalized task setting aligned with emerging technical trends. 
Traditional VFI focuses primarily on synthesizing a fixed, limited number of intermediate frames (typically fewer than 25) between adjacent frames with minimal visual differences, offering no control over the generated content. However, our SFI task accepts two potentially dissimilar input frames and optional textual guidance to generate intermediate sequences of arbitrary length. This advanced formulation enables the handling of substantially different input frames while supporting precise control through text prompts and accommodating unrestricted frame count requirements. 
The combination of conditional frame \textbf{semantic extraction}, contextual (video-text) \textbf{semantic fusion}, and frame adaptive \textbf{semantic reasoning} (texture understanding for fewer frames and subject semantic understanding for longer frames) collectively forms the \textbf{SEMANTIC} completion for the frame interpolation task.
Moreover, while VFI is primarily limited to applications such as slow motion generation, frame rate improvement, and frame recovery, our SFI framework extends these capabilities to enable advanced applications, including intelligent scene transitions, special effects blending, and creative video generation with semantic level control, which brings a significant expansion of the practical utility for frame interpolation technologies in professional video production and digital content creation.
Second, at the methodology level, we develop \textbf{SemFi}, a novel end-to-end framework that innovatively adapts FVM architectures for SFI. We augment the explicit and implicit information injected into the model to achieve control over the content of the first and last frames. Meanwhile, recognizing the critical need for frame-adaptive processing, we introduce a dynamic Mixture-of-LoRA (MoL) architecture where specialized adapters are automatically selected based on target frame counts, enabling optimized performance across different interpolation intervals. As shown in Fig.~\ref{fig:1}, this integrated approach enables high-quality interpolation results regardless of output sequence length while maintaining strict adherence to user specifications. Finally, to support the ambitious task, we construct \textbf{SFI-300K}, a large-scale multimodal dataset featuring 300,000 high-quality video clips spanning diverse content categories, with frame intervals ranging from 5 to 81 frames.  Each sample includes rich and high-density captions. Building upon this foundation, we introduce SFIBench, the first standardized evaluation framework for SFI that assesses model performance in multiple dimensions, including temporal consistency, visual quality, and instruction adherence at various generation lengths.

In summary, our contributions are threefold:
\begin{itemize}
  \item We formally define semantic frame interpolation as a novel generative inbetweening paradigm that enables customized intermediate content generation under more challenging control conditions, significantly expanding the diversity of producible outputs. 
  \item We propose SemFi, a novel framework for SFI, introducing Mixture-of-LoRA for adaptive generation. It achieves better performance across a wide range of frame counts while maintaining precise control.
  \item We introduce SFI-300K, the first large-scale dataset for SFI, featuring 300K diverse clips with rich annotations. The accompanying SFIBench provides standardized evaluation across fidelity, coherence, and instruction adherence at varying generation lengths.
\end{itemize}

\section{Related Work} \label{sec:related_work}

\textbf{Video Frame Interpolation.} Video Frame Interpolation (VFI)~\cite{dong2023video,kiefhaber2024benchmarking} generates intermediate frames in a video sequence to boost frame rate and visual smoothness, with methods mainly categorized into kernel-based, optical flow-based, and diffusion-based. Kernel-based methods~\cite{cheng2021multiple} convolve source frames with adaptive~\cite{briedis2023kernel,niklaus2017video}, deformable~\cite{tian2022video,xiang2020zooming}, or attention map kernels~\cite{lu2022video,shi2022video} to create intermediate frames, performing well in simple scenarios~\cite{choi2020channel} but limited by kernel size and interpolation positions in complex motion cases~\cite{niklaus2017video2}.
Optical flow-based methods~\cite{ding2024video,prasanna2024video} use optical flows for interpolation, which estimate flow between frames and apply warping techniques~\cite{niklaus2020softmax,shimizu2022forward}, with some methods incorporating flow reversal or predicting intermediate flows~\cite{niklaus2020softmax,zhang2023extracting,seo2024bim}. They have high computational complexity and perform poorly in complex scenarios~\cite{park2020bmbc}. With diffusion models' development~\cite{ho2022video,yang2023diffusion}, diffusion-based VFI models emerged~\cite{danier2024ldmvfi,huang2024motion, zhu2024generative}. Strategies include training with large-scale data~\cite{hur2025high,jain2024video,xing2024dynamicrafter} and using pre-trained models with novel sampling techniques, such as TRF~\cite{feng2024explorative} and VIBIDSampler~\cite{yang2025vibidsampler}, to enhance both quality and efficiency. In general, kernel-based and optical flow-based methods struggle in complex scenarios, and existing diffusion-based methods are motion-dependent, limiting interpolation diversity. To solve these problems, we introduce the SFI task, aiming for more diverse and accurate interpolation.

\textbf{Video Generation Models.} Early video diffusion models like Stable Video Diffusion~\cite{blattmann2023stable}, AnimateDiff~\cite{guo2023animatediff} extend pretrained 2D UNets for images. They add extra modules to manage temporal attention, implementing a 2D+1D attention mechanism within the 2D latent space of a pretrained 2D VAE. Later, Sora~\cite{brooks2024video} plays a crucial role in this development by demonstrating the scalability and numerous benefits of the diffusion transformer (DiT) architecture. Subsequently, recent works like CogVideoX~\cite{yang2024cogvideox}, Hunyuan Video~\cite{kong2024hunyuanvideo}, and Wan~\cite{wang2025wan}, which adopt the DiT architecture, have achieved outstanding results, further validating the effectiveness of this approach.
Despite these advances, current video generation research mainly focuses on Text-to-Video (T2V), or Image-to-Video (I2V) with a single reference. Frame interpolation between the first and last frames, a practical and important function, remains underexplored. Existing solutions, whether open-source like Wan~\cite{wang2025wan} or closed-source like Kling~\cite{kelin}, struggle with large frame differences and have limitations in the scale of generated frames. In contrast, our model effectively handles complex, changing scenarios and flexibly addresses multi-scale issues, enabling high-quality frame interpolation with enhanced adaptability and precision.

\textbf{Mixture-of-Experts and Adapter-Based Specialization.} Mixture-of-Experts (MoE) methods improve model capacity by routing different inputs to sparse experts~\cite{jacobs1991adaptive, shazeer2017outrageously, fedus2022switch}. Adapter-based methods instead introduce lightweight modular adaptation for efficient transfer and composition across tasks~\cite{pfeiffer2021adapterfusion}. Related ideas have also been explored in parameter-efficient tuning, where multiple LoRA modules are composed, fused, or dynamically selected to improve adaptation flexibility~\cite{hu2022lora, huang2023lorahub}. More recent works further extend this line by introducing explicit mixtures of LoRA experts or dynamic LoRA routing~\cite{wu2024mixture, kong2024lora}. Different from these methods, our proposed Mixture-of-Lora is not designed for generic task routing, semantic-domain specialization, or token-wise expert selection. Instead, it is introduced for the temporal-scale heterogeneity in SFI: the routing variable is the target frame count, and each expert is anchored to a local temporal regime for variable-length semantic interpolation.

\section{Methodology: Formulation, Architecture and Dataset} \label{sec:method}

This section formally establishes the conceptual foundation of Semantic Frame Interpolation (SFI) by presenting its precise definition and technical formulation. Building upon this theoretical framework, we subsequently introduce SemFi, our novel neural architecture that synergistically combines foundation model priors with adaptive low-rank adaptations for controllable intermediate frame generation. The section culminates with the presentation of SFI-300K, the first comprehensive benchmark dataset specifically designed for SFI tasks, which establishes standardized evaluation protocols, including rich and diverse assessment metrics. Our methodological pipeline progresses from theoretical formulation (Section~\ref{sfi}) to architectural innovation (Section~\ref{semfi}) and finally to benchmark establishment (Section~\ref{sfi-300k}), providing complete coverage of both algorithmic and evaluative dimensions for advancing SFI research.

\subsection{Task Formulation: Semantic Frame Interpolation} \label{sfi}

\textbf{Semantic frame interpolation} aims to generate intermediate frames between two endpoint images under both frame-count and semantic control. Compared with conventional VFI, which mainly addresses short-range interpolation without semantic guidance, SFI considers more general settings where the start and end frames may differ substantially and the desired number of intermediate frames can vary. An optional text prompt further specifies the intended transition behavior.

Formally, let $I_f \in \mathbb{R}^{H \times W \times C}$ and $I_l \in \mathbb{R}^{H \times W \times C}$ denote the first and last frames, $T$ denote an optional text prompt and $N \in \mathbb{Z}^+$ denote the target number of intermediate frames. The goal of SFI $\mathcal{F}$ is to generate a sequence $\{I_1, I_2, ..., I_N\}$ that is visually consistent with both endpoints and semantically aligned with the prompt. This can be written as:
\begin{align*}
    \{I_1, I_2, ..., I_N\} = \mathcal{F}(I_f, I_l, T, N).
\end{align*}
More generally, SFI can be viewed as a conditional generation problem with distribution:
\begin{align*}
    p(I_{1:N}|I_f, I_l, T, N),
\end{align*}
where the model aims to generate the most plausible intermediate sequence under the given conditions:
\begin{equation}
\{I_1, I_2, ..., I_N\} = \arg \max_{I_{1:N}} p_\theta(I_{1:N}|I_f, I_l, T, N).
\end{equation}
When $N$ is small and $T = \emptyset$, SFI reduces to conventional video frame interpolation; when $N$ is large, it approaches foundation-model-based frame-to-frame generation. Therefore, SFI provides a unified formulation for controllable interpolation across different temporal scales.

\subsection{SemFi: Mixture-of-LoRA for Semantic Multi-frame Interpolation}
\label{semfi}

\begin{figure*}[tbp]
    \centering
    \includegraphics[width=\linewidth]{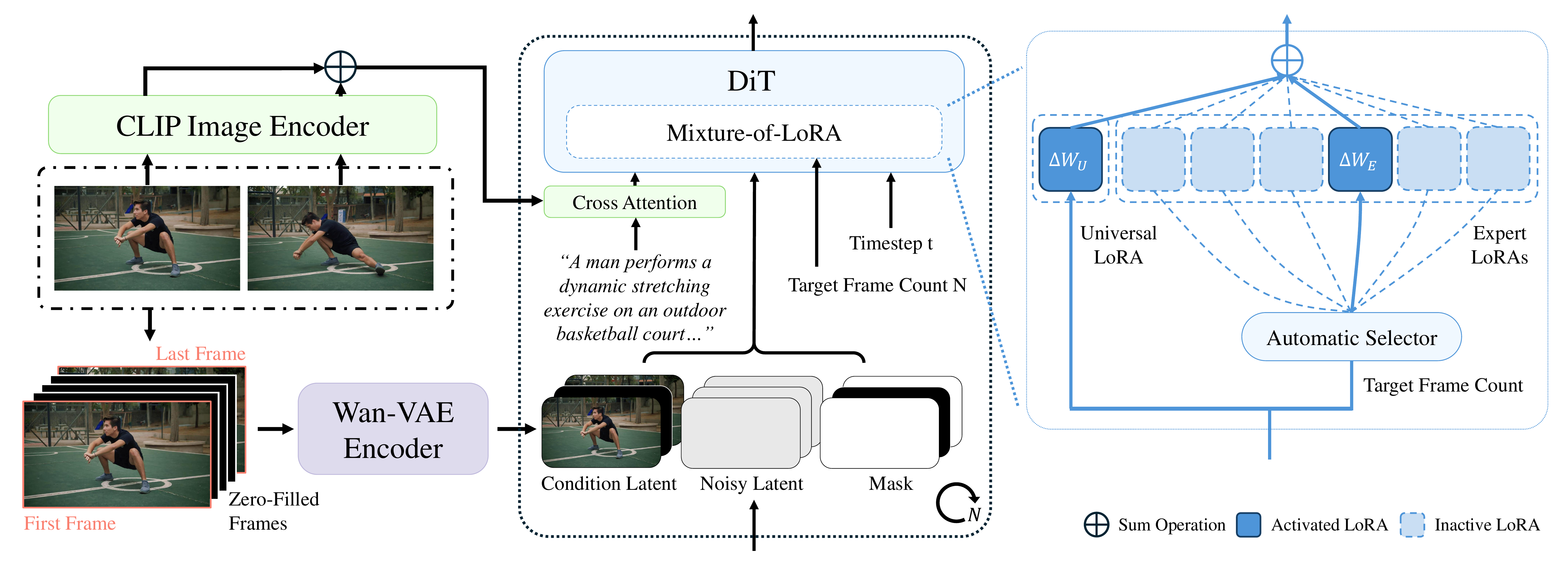}
    \caption{\textbf{Overview architecture of the SemFi model.} SemFi employs a mixture-of-LoRA module to dynamically activate the most suitable LoRA parameters for the target frame count, enabling high-quality semantic frame interpolation across diverse frame generation requirements.}
    \label{fig:2}
\end{figure*}

As shown in Fig.~\ref{fig:2}, building upon the formal definition of SFI, we propose SemFi, a novel architecture constructed upon the 14B-parameter Wan2.1 I2V foundation model \cite{wang2025wan}. Our design harnesses the powerful priors of FVMs while introducing precise semantic control over intermediate frame generation. Besides, to achieve robust performance across varying frame counts, SemFi incorporates a Mixture-of-LoRA (MoL) mechanism that dynamically activates specialized adapters based on target frame requirements, enabling unified ``interpolation", ``transition", and ``generation".

\subsubsection{Conditional Information Injection}

Formulated through the lens of FVMs, SFI is fundamentally a more strictly constrained I2V generation task where both the start and end frames serve as boundary conditions, which inspires us to adopt the Wan-I2V architecture as our backbone. Unlike text-to-video models, I2V additionally processes a concatenated tensor $I_c$, the guidance frames, along the temporal dimension, consisting of the condition image (the initial frame) and zero-filled subsequent frames, which is then encoded by Wan-VAE into latent representations. Meanwhile, Wan-I2V introduces a binary mask $M$ distinguishing preserved (1) and generated (0) frames. After projection, $I_c$, $M$ and the noisy latent are channel-concatenated for Wan-DiT processing. Moreover, Wan-I2V utilizes the image encoder of CLIP to extract the feature representation of the condition image and injects it into DiT through cross-attention after projection. To extend the original single-image conditioning to dual-frame control while maintaining compatibility with the Wan-I2V architecture, we implement three key modifications to the feature processing pipeline: First, the guidance frame construction is reformulated by setting the initial frame $I_f$ and the last frame $I_l$ as the first and last elements of the temporal sequence respectively, with zero-padded intermediate frames occupying positions $1$ to $N-2$. Second, the binary mask $M$ is correspondingly configured with $M_{0} = M_{N-1} = 1$ and $M_{1:N-2} = 0$. Finally, CLIP image embeddings from both condition frames are summed before projection, maintaining the original cross-attention injection mechanism. These modifications help SemFi maintain Wan's text-conditioning capability while enabling dual endpoint control.

\subsubsection{Mixture-of-LoRA for Multi-Frame Generation}

Fine-tuning SemFi by training LoRA is intuitive and efficient. However, our preliminary experiments with single-LoRA adaptation on the multi-frame scale SFI-300K dataset revealed a fundamental limitation: a unified low-rank adaptation matrix struggles to capture the distinct feature distributions required for different interpolation scales. This motivates our proposed mixture-of-LoRA architecture, which introduces specialized expert generation while preserving shared knowledge transfer. The MoL employs 1 universal LoRA and 6 frame-count-specific LoRA experts. The \textbf{universal LoRA} $\Delta W_U$ is a base adapter trained on all frame counts that captures interpolation-invariant features, including motion consistency and semantic preservation across scales. This module remains active throughout all training and inference phases. The \textbf{expert LoRAs} $\{ \Delta W_{E_s}\}_{s \in S}$ are frame-specific adapters specializing in six different discrete frame counts $S$. For any forward process, exactly one expert LoRA is selectively activated, which exclusively specializes in enhancing the interpolation characteristics of a certain frame count. For target frame counts outside the trained discrete scales $S$, SemFi automatically selects the most appropriate expert LoRA by minimizing the absolute temporal distance $|N - s|$ where $s \in S$, ensuring graceful generalization to unseen interpolation requirements. The complete adaptation is:
\begin{equation}
\Delta W = \Delta W_U + \Delta W_{E_s}, ~~~ s = \arg\min_{s \in S} |N - s|.
\end{equation}
We define the frame count set for expert LoRAs as $S = \{5, 9, 17, 33, 65, 81\}$, which is selected according to both task coverage and structural efficiency. From the task perspective, these anchors are designed to cover different interpolation demands ranging from short-range interpolation to long-term semantic generation. Specifically, the 5-frame and 9-frame settings target conventional interpolation scenarios that mainly require fine-grained motion continuity; the 17-frame and 33-frame settings are suitable for medium-length transitions with moderate temporal variation; and the 65-frame and 81-frame settings are intended for long-range generation, where semantic bridging between distant endpoints becomes dominant. In this way, the selected anchors provide coverage over qualitatively different temporal regimes in SFI. From the structural perspective, let $N_{max}$ denote the maximum frame count that can be stably trained under the current computational resources. We construct the anchor set in the form:
\begin{align}
\label{eq:1}
    S = \{2^n + 1 | n \in \mathbb{N}, n \geq 2, 2^n+1 < N_{max}\} \cup \{N_{max}\}.
\end{align}
Under our current setting, this yields $\{5, 9, 17, 33, 65, 81\}$. Such a geometric progression ensures that, when the trainable frame range expands, i.e., $N_{max}$ increases, the number of experts $|S|$ does not grow excessively and cause prohibitive computational overhead. Therefore, the frame-count set $S$ is essentially a scalable partition of the currently trainable temporal range. When the upper bound of trainable frame counts increases, the anchor set can be correspondingly extended while preserving the same design principle. In this way, the selected set achieves a practical balance among temporal coverage, expert specialization, and computational cost.

The role of MoL is to allow the model to respond differently to different target frame counts while preserving shared interpolation knowledge. In practice, the universal LoRA provides a common adaptation basis across all cases, while the selected expert LoRA further adjusts the generation behavior for the corresponding temporal scale. Unlike traditional feature-based interpolation methods that rely on generic features to cover all interpolation cases, MoL explicitly refines the functional roles of learned features in SemFi. It enables scale-invariant knowledge, such as general consistency priors, and scale-dependent knowledge, such as local continuity for short-range interpolation and semantic bridging for long-range generation, to be modeled separately. In this way, MoL not only enhances feature adaptation, but also serves as an SFI-oriented mechanism for organizing and activating different interpolation knowledge according to the target temporal regime.

In addition, MoL also exhibits transferability. The proposed MoL strategy can be applied to a broad class of image-to-video generation models. In practice, it only requires that the model be able to inject and process image-conditioned features, and that its backbone contain Transformer blocks or other modules where low-rank residual adapters can be inserted. Under these conditions, a universal LoRA can be used to capture interpolation priors shared across frame scales, while expert LoRAs provide additional adaptation for different temporal regimes. This makes MoL a practical way to extend existing image-conditioned video generation frameworks toward variable-length semantic interpolation.

\subsection{SFI-300K: A High-Quality and Large-Scale SFI Dataset}
\label{sfi-300k}

\begin{table*}[tbp]
    \centering
    \caption{Comparison with popular frame interpolation datasets on different attributes.}
    \label{tab:comparison_dataset}
    \begin{tabular}{ccccccccccc}
    \toprule
    \multirow{2}{*}{Dataset} & \multirow{2}{*}{\makecell[c]{Year}} & \multirow{2}{*}{\makecell[c]{Total \\ Minutes}} & \multirow{2}{*}{\makecell[c]{Total \\ Videos}} & \multirow{2}{*}{\makecell[c]{Source \\ Resolution}} & \multirow{2}{*}{\makecell[c]{Frames}} & \multirow{2}{*}{\makecell[c]{Semantic\\Information}} & \multirow{2}{*}{\makecell[c]{Task \\ Domain}} \\ 
    \\
    \hline
    UCF101~\cite{soomro2012ucf101} & 2012 & 1600 & 13320 & 320$\times$240 & 29$\sim$1776 & \xmark & VFI \\
    Vimeo90K~\cite{xue2019video} & 2019 & - & 73,171 & 448$\times$256 & 3 & \xmark & VFI\\
    Xiph~\cite{niklaus2020softmax} & 2020 & 4 & 19 & 4096$\times$2160 & 255$\sim$1200 & \xmark & VFI \\ 
    Inter4K~\cite{stergiou2022adapool} & 2021 & 83 & 1,000 & 3840$\times$2160 & 300 & \xmark & VFI\\
    GI Datasets~\cite{wang2024generative} & 2024 & 8 & 113 & 1024$\times$576 & 97 & \xmark & VFI\\
    \hline
    \rowcolor{blue_tab}
    SFI-300K & 2025 & 5833 & 300,000 & 1920$\times$1080, 1080$\times$1920, 2048$\times$1080,... & 5$\sim$81 & \cmark & \textbf{SFI}\\
    \bottomrule
    \end{tabular}
\end{table*}

\begin{figure*}[tbp]
    \centering
    \includegraphics[width=\linewidth]{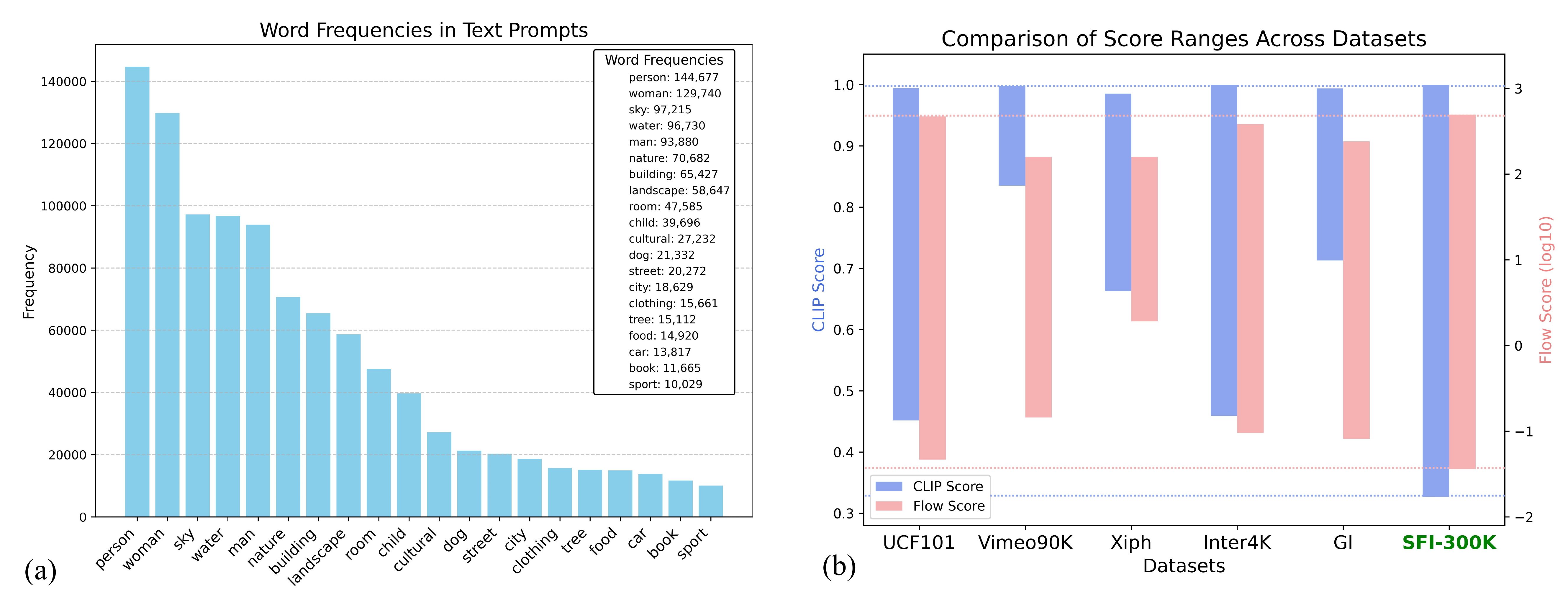}
    \caption{Dataset analysis. (a) SFI-300K demonstrates substantial diversity across video categories; (b) SFI-300K exhibits a broader distribution of CLIP scores and flow scores, indicating richer variation in content and motion between the first and last frames.}
    \label{fig:3}
\end{figure*}

\subsubsection{Data Collection and Annotation}

As shown in Table~\ref{tab:comparison_dataset}, previous datasets related to SFI have several key limitations. The datasets employed by SOTA methods in frame interpolation are either relatively small in scale, lack semantic cues, or consist of short ground-truth videos that contain few frames, often with minimal differences between the first and last frames. While recent advancements in large video models have led to the release of high-quality, large-scale video datasets \cite{lin2024open, zhang2026ultravideo}, these are more general in nature and are not specifically designed for SFI tasks. As a result, they fall short in meeting the unique requirements for model training and evaluation. To address these gaps, we curate a dataset specifically tailored for the SFI task, building upon the publicly available datasets used in Open Sora Plan \cite{lin2024open}. Our dataset is carefully constructed by filtering based on the degree of difference between the first and last frames, segmenting multi-frame clips, and annotating high-quality captions. 

To standardize the data and ensure the effectiveness of subsequent operations, we first filter out videos $\mathcal{V}_{\text{filtered}}$ with a frame rate less than or equal to 30 and a total frame count between $f_{\text{max}}$ and $4f_\text{max}$, where $f_{\text{max}}$ is the maximum frame count in our multi-frame scale dataset, set to 81 frames. Next, we extract the first and last frames of each video and use the CLIP \cite{radford2021learning} and RAFT \cite{teed2020raft} methods to compute the CLIP score $S_c$ and flow score $S_f$. The CLIP score quantifies the similarity between the image features of the first and last frames, with a higher $S_c$ indicating greater similarity. The flow score \cite{wang2024koala} represents the average optical flow $L_2$-norm across all pixels, with a higher $S_f$ indicating greater differences between the first and last frames. By analyzing the CLIP and flow scores across all videos, we define high and low thresholds for each score and remove any videos that fall outside the threshold range. This ensures that the final dataset contains only videos with meaningful changes between the first and last frames. Following this, we perform multi-frame trimming of the videos according to the frame count set $S$ mentioned in Section~\ref{semfi}. The final ground truth videos $\mathcal{V}$ are generated through the segmentation process:
\begin{equation}
\mathcal{V} = \bigcup_v \text{Cut}(v, \frac{f}{2} - \frac{s}{2}, \frac{f}{2} + \frac{s}{2}), \quad s \in S = \{5, 9, 17, 33, 65, 81\},
\end{equation}
where $v \in \mathcal{V}_{\text{filtered}}$ represents a video to be segmented, $\text{Cut}(v, x, y)$ refers to extracting the clip from the $x$-th frame to the $y$-th frame from video $v$, and $f$ is the number of frames in video $v$. We then use the Qwen2.5-VL-32B model \cite{Qwen-VL, Qwen2.5-VL} and a carefully designed prompt to annotate all clips with high-perception, high-information-density long-text semantic captions $\mathcal{C}$, finally resulting in the dataset SFI-300K $\mathcal{D} = \{\mathcal{V}, \mathcal{C}\}$, containing 300,000 video clips. Table~\ref{tab:comparison_dataset} and Fig.~\ref{fig:3} show SFI-300K's better alignment with SFI task requirements than other datasets.

\subsubsection{SFIBench}
\label{sfibench}

We generate an additional 400 video clips with the same data processing pipeline, forming a general-purpose test set used for evaluating SFI models in SFIBench. However, some key differences need to be noted: 1) The frame counts in the test set are not limited to the six frame counts in $S$, but rather span a range from $3$ frames to $102$ frames, covering $100$ different frame counts in total. Each frame count has $4$ video clips, resulting in a total of $400$ video clips. Expanding the frame count range allows for a more comprehensive evaluation of the model’s performance, rather than focusing solely on generation results for certain frame counts. 2) Unlike the training set, where each video generates $6$ different frame-count clips, in the test set, each original video is used to generate only a single video clip. This approach enhances the diversity of the test data.

In addition, the evaluation of the SFI task requires specialized quantitative metrics. Traditional VFI or video generation benchmarks \cite{soomro2012ucf101, stergiou2022adapool, wang2025framer, ivebench, humanvideo} typically assess performance by computing the fidelity between each generated frame and its corresponding ground-truth frame. However, in the SFI task, the goal of interpolation should not be defined as the exact reproduction of the ground truth. Instead, SFI emphasizes content generation, where the key concerns are whether the generated intermediate frames, as a complete video, exhibit perceptually reasonable qualities and whether they align with the given textual prompts. Accordingly, in SFIBench, we evaluate SFI performance along two critical dimensions: video quality and semantic consistency. First, to monitor the perceptual quality of the generated videos, we adopt the one-for-all multi-dimensional quality assessment paradigm from FineVQ \cite{duan2025finevq}, conducting systematic quantitative analysis across six dimensions: color, noise, artifact, blur, temporal consistency, and overall quality. Second, to measure semantic alignment between the generated results and the textual prompts, we leverage state-of-the-art multimodal large language models. We design a carefully crafted, multi-layered text-consistency evaluation prompt, and employ both GLM-4.5V \cite{hong2025glm} and Qwen3-VL-235B-A22B-Instruct \cite{Qwen-VL, yang2025qwen3} to assign semantic consistency scores to the outputs.
\begin{table*}[tbp]
\renewcommand{\arraystretch}{1.2}
\centering
\caption{Quantitative comparison. We highlight the \colorbox[HTML]{ffaeb0}{best}, \colorbox[HTML]{ffd7ac}{second-best}, and \colorbox[HTML]{ffffa8}{third-best} results for each metric. }
\begin{tabular}{lcccccccc}
\hline
        & \multicolumn{6}{c}{Video Quality} & \multicolumn{2}{c}{Semantic Consistency} \\ \cline{2-9}
Methods & Color↑ & Noise↑ & Artifact↑ & Blur↑ & Temporal↑ & Overall↑ & GLM-4.5V↑ & Qwen3-VL↑ \\ \hline
Framer  & 39.25 & 40.39 & 33.52 & 33.12 & 43.21 & 35.30 & 83.80 & 91.90 \\
GI      & 42.17 & 42.85 & 37.45 & 37.02 & 43.85 & 38.81 & 81.90 & 86.53 \\
FCVG    & \colorbox[HTML]{ffd7ac}{46.43} & \colorbox[HTML]{ffd7ac}{46.90} & \colorbox[HTML]{ffd7ac}{41.76} & \colorbox[HTML]{ffd7ac}{41.38} & \colorbox[HTML]{ffd7ac}{48.00} & \colorbox[HTML]{ffd7ac}{43.46} & \colorbox[HTML]{ffffa8}{85.60} & \colorbox[HTML]{ffffa8}{93.66} \\
Wan     & \colorbox[HTML]{ffffa8}{43.90} & \colorbox[HTML]{ffffa8}{44.67} & \colorbox[HTML]{ffffa8}{38.91} & \colorbox[HTML]{ffffa8}{38.46} & \colorbox[HTML]{ffffa8}{46.93} & \colorbox[HTML]{ffffa8}{40.62} & \colorbox[HTML]{ffd7ac}{86.14} & \colorbox[HTML]{ffd7ac}{94.08} \\ \hline
SemFi   & \colorbox[HTML]{ffaeb0}{49.65} & \colorbox[HTML]{ffaeb0}{50.37} & \colorbox[HTML]{ffaeb0}{46.33} & \colorbox[HTML]{ffaeb0}{46.00} & \colorbox[HTML]{ffaeb0}{52.49} & \colorbox[HTML]{ffaeb0}{47.95} & \colorbox[HTML]{ffaeb0}{86.36} & \colorbox[HTML]{ffaeb0}{94.39} \\ \hline
\end{tabular}
\label{tab:2}
\end{table*}

\begin{figure*}[tbp]
    \centering
    \includegraphics[width=\linewidth]{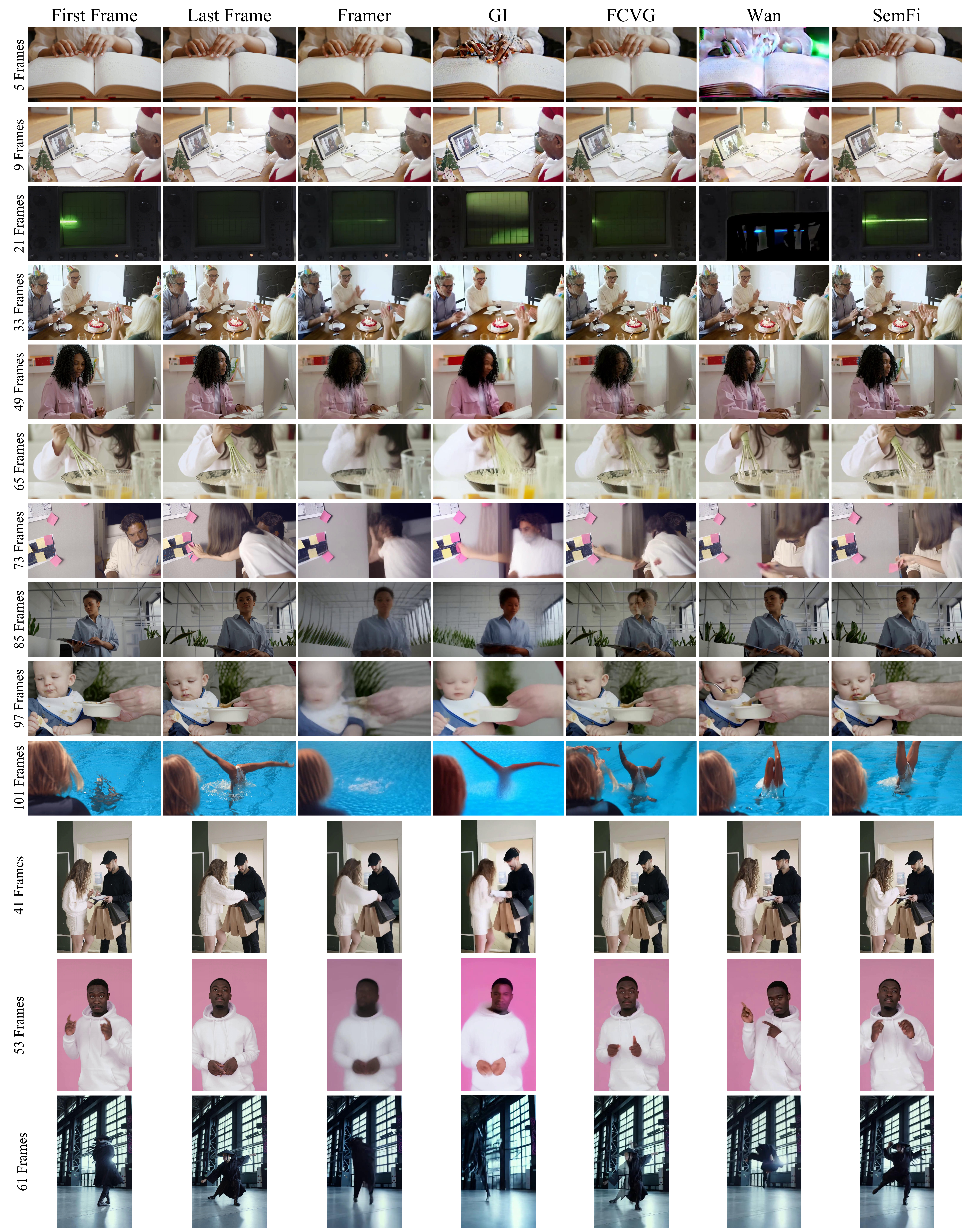}
    \caption{Qualitative comparison with Framer~\cite{wang2025framer}, GI~\cite{wang2024generative}, FCVG~\cite{zhu2024generative} and Wan2.1~\cite{wang2025wan}. Wan produces noticeable artifacts at lower frame counts and tends to generate exaggerated, unnatural motions in high-magnitude scenarios. Other VFI methods often fail in long-range interpolation, resulting in complete distortion or oscillatory patterns between reference frames. In contrast, our SemFi consistently produces coherent and realistic interpolation results across varying frame counts and resolutions.}
    \label{fig:4}
\end{figure*}

\section{Experiments} \label{sec:exp}

\subsection{Implementation Details}

We adapt the Wan-14B-I2V foundation model \cite{wang2025wan} to construct SemFi through a two-stage training strategy with parameter-efficient fine-tuning. In the first stage, we freeze all original model parameters and exclusively train the universal LoRA on a randomly selected half of the SFI-300K. In the second stage, we freeze both the base model parameters and the universal LoRA parameters, and train 6 expert LoRAs on the remaining portion of the training dataset. All LoRAs are configured with a rank of 16. Both fine-tuning processes are conducted for 1 epoch on their respective training data using the AdamW optimizer on 32 NVIDIA H20 GPUs, with a global batch size of 32 and a learning rate of $1 \times 10 ^{-4}$. Consistent with Wan's default configuration, all videos are resized to $480 \times 832$ resolution during training.

\subsection{Comparison with State-of-the-Arts}

\textbf{Baselines.} We compare SemFi with 4 other models capable of semantic frame interpolation. Framer \cite{wang2025framer}, GI \cite{wang2024generative} and FCVG \cite{zhu2024generative} are VFI models. They are essentially diffusion-based video generation networks and can be adapted to support SFI. Wan \cite{wang2025wan}, a foundation video model, also features frame-to-frame capabilities, enabling intermediate frame generation given the first and last frames.

\textbf{Quantitative Comparison.} Since both Wan and SemFi only support generating videos with frame counts of the form $4n + 1, n \in \mathbb{N}$, we select all corresponding cases from SFIBench for evaluation. For each baseline, we use its officially released code and generate results for all test cases under the default inference settings provided by the corresponding implementation. As described in Section~\ref{sfibench}, we compute 8 metrics to assess model performance. For the 6 video-quality dimensions, namely Color, Noise, Artifact, Blur, Temporal, and Overall, we directly adopt the default score inference scripts released in the open-source FineVQ repository and evaluate all generated videos under the same configuration. For text consistency evaluation, we interact with GLM-4.5V and Qwen3-VL using a carefully designed evaluation prompt. For each generated video, the model output is truncated with a maximum generation budget of 8192 tokens, and a numerical score in the range of 0 to 100 is extracted from the response to reflect the consistency between the generated content and the corresponding text description. The final evaluation results are presented in Table~\ref{tab:2}, which demonstrate that SemFi outperforms all baselines. In particular, it achieves significant improvements in video quality, highlighting its ability to produce high-quality and stable interpolation results across a wide range of frame counts.

\begin{table*}[tbp]
\centering
\caption{Quantitative comparison on conventional VFI benchmarks. Best results are shown in \textbf{bold} and second-best results are \underline{underlined}.}
\label{tab:4}
\begin{tabular}{lcccccccccc}
\hline
 & \multicolumn{5}{c}{\textbf{DAVIS-7}} & \multicolumn{5}{c}{\textbf{UCF101-7}} \\ \cline{2-11} 
 & PSNR↑ & SSIM↑ & LPIPS↓ & FID↓ & \multicolumn{1}{c|}{FVD↓} & PSNR↑ & SSIM↑ & LPIPS↓ & FID↓ & FVD↓ \\ \hline
Framer & \textbf{19.58} & \textbf{0.5184} & 0.2530 & \underline{30.15} & \multicolumn{1}{c|}{\underline{589.17}} & \textbf{23.18} & \textbf{0.8514} & \underline{0.0900} & \underline{21.91} & \textbf{539.22} \\
GI     & 12.15 & 0.2434 & 0.5076 & 147.3 & \multicolumn{1}{c|}{1940.1} & 11.96 & 0.4060 & 0.4165 & 150.5 & 4488.8 \\
FCVG   & 18.35 & 0.4485 & \underline{0.2327} & 31.37 & \multicolumn{1}{c|}{686.05} & 18.51 & 0.6413 & 0.1415 & 29.61 & 659.33 \\
Wan & 14.41 & 0.3654 & 0.3558 & 73.85 & \multicolumn{1}{c|}{1234.1} & 7.491 & 0.2213 & 0.7045 & 293.9 & 7387.2 \\ \hline
SemFi & \underline{19.39} & \underline{0.4949} & \textbf{0.1838} & \textbf{21.34} & \multicolumn{1}{c|}{\textbf{541.87}} & \underline{22.03} & \underline{0.7814} & \textbf{0.0749} & \textbf{14.46} & \underline{600.35} \\ \hline
\end{tabular}
\end{table*}

\textbf{Qualitative Comparison.} Given that SFI is inherently a generative task, it should not be evaluated solely based on quantitative metrics. Therefore, we conduct a qualitative comparison to enable a comprehensive and fair evaluation of all models. As shown in Fig.~\ref{fig:4}, when generating a small number of frames, Wan's results exhibit significant artifacts, distortions, blurring, and chromatic aberrations, failing to produce acceptable outputs. As the frame count increases, the performance of VFI methods (Framer, GI, and FCVG) progressively deteriorates. With larger frame counts and greater disparities between start and end frames, these VFI approaches struggle to synthesize sufficient and natural transitional content. In most cases, they simply flash or fade abruptly from the first to the last frame. In contrast, our SemFi consistently generates reasonable, high-quality intermediate frames regardless of frame count requirements. It effectively interprets users' specific interpolation needs by analyzing both the disparity between conditional frames and the requested frame count.

\subsection{Evaluation under Conventional VFI Settings}

To further verify that the proposed SFI formulation indeed covers conventional video frame interpolation when the target frame count is small, and no text control is required, we additionally evaluate SemFi under standard VFI settings. Specifically, following the benchmark protocol proposed in VIDIM~\cite{jain2024video}, we construct evaluation subsets from DAVIS~\cite{pont20172017} and UCF101~\cite{soomro2012ucf101} using its official test-data generation procedure, resulting in 357 video clips for DAVIS and 400 for UCF101, which cover a mixture of both small- and large-motion scenarios. Each clip contains 9 frames, where the first and last frames are used as model inputs and the middle 7 frames are treated as the ground-truth interpolation targets. For SemFi and Wan, we use the first and last frames as conditional inputs and set the text prompt to be empty, so that the evaluation reduces to a conventional interpolation setting. Following VIDIM, we report standard quantitative metrics including PSNR, SSIM~\cite{wang2004image}, LPIPS~\cite{zhang2018unreasonable}, FID~\cite{heusel2017gans}, and FVD~\cite{unterthiner2018towards}, and compare SemFi and Wan with representative recent VFI methods. For each method, we first resize the input frames while preserving the original aspect ratio to an inference shape close to the training-resolution range of the corresponding method. After inference, the generated results are resampled back to the original benchmark resolution for metric computation. The quantitative comparison is shown in Table~\ref{tab:4}. As expected, the frame-to-frame generation model Wan performs poorly under conventional VFI settings. In contrast, SemFi, although trained for the broader SFI task, remains highly competitive under standard VFI evaluation. It is worth noting that VIDIM also discusses that reconstruction-oriented metrics such as PSNR, SSIM, and LPIPS may penalize plausible but different extrapolated results, and therefore may not fully reflect the quality of generative interpolation models. In particular, PSNR and SSIM favor pixel-level alignment with a single ground-truth trajectory, while generative models may synthesize visually reasonable intermediate results that deviate from the exact reference frames. Therefore, we interpret the conventional VFI results from both reconstruction-oriented metrics and generation-oriented metrics. While Framer achieves the best PSNR and SSIM, SemFi performs particularly well on FID and FVD, which are commonly used to evaluate the distributional quality of generated visual content.

Overall, these results provide additional evidence that the proposed SFI formulation indeed includes conventional VFI as a special case. At the same time, we acknowledge that SemFi is designed for a broader interpolation setting involving semantic control and variable-length generation, rather than being exclusively optimized for short-range reconstruction under a specific low-resolution VFI benchmark. Therefore, SemFi may not always surpass methods specifically trained for this narrow regime on every reconstruction metric. Nevertheless, under the broader SFI framework, SemFi shows clear advantages over previous baselines while maintaining competitive performance on conventional VFI tasks.

\subsection{Comparison with Conventional 2$\times$ VFI Methods}

To further evaluate the effectiveness of the proposed SFI formulation under long-sequence interpolation, we additionally compare SemFi with conventional recursive 2$\times$ VFI methods. Specifically, we adopt FILM~\cite{reda2022film}, RIFE~\cite{huang2022real}, and EMA-VFI~\cite{zhang2023extracting} as representative 2$\times$ interpolation baselines. Note that for EMA-VFI, we evaluate it by recursively applying its 2$\times$ interpolation pipeline, following the same protocol as FILM and RIFE. Since recursive 2$\times$ interpolation naturally supports frame counts in the form of $2^n+1$, we evaluate two long-sequence settings, 65 frames and 129 frames, with 30 cases for each frame count. Both settings are substantially longer than conventional short-range VFI. Moreover, the 129-frame setting exceeds the maximum training frame count of SemFi, i.e., 81 frames, by a large margin, making it a stricter test of long-sequence generalization. We evaluate all methods using the same video-quality and semantic-consistency metrics as in SFIBench. The quantitative results are reported in Table~\ref{tab:8}. SemFi achieves the best performance across all video-quality dimensions and semantic consistency metrics. These results indicate that directly modeling the target frame count is more effective for long-sequence interpolation than recursively applying a short-range 2$\times$ interpolator.

\begin{table*}[tbp]
\small
\centering
\caption{Quantitative comparison with conventional 2$\times$ VFI methods in long-sequence interpolation. Best results are shown in \textbf{bold}.}
\begin{tabular}{lcccccccc}
\hline
        & \multicolumn{6}{c}{Video Quality} & \multicolumn{2}{c}{Semantic Consistency} \\ \cline{2-9}
Methods & Color↑ & Noise↑ & Artifact↑ & Blur↑ & Temporal↑ & Overall↑ & GLM-4.5V↑ & Qwen3-VL↑ \\ \hline
FILM    & 48.24 & 49.11 & 47.90 & 47.65 & 49.72 & 48.42 & 78.50 & 82.00 \\
RIFE    & 47.02 & 47.02 & 47.70 & 47.41 & 47.95 & 47.68 & 77.08 & 79.42 \\
EMA-VFI & 45.20 & 45.34 & 45.18 & 44.68 & 45.67 & 45.17 & 72.92 & 80.25 \\
SemFi   & \textbf{49.96} & \textbf{49.72} & \textbf{50.72} & \textbf{50.59} & \textbf{50.44} & \textbf{50.68} & \textbf{80.92} & \textbf{85.67} \\ \hline
\end{tabular}
\label{tab:8}
\end{table*}

\begin{figure*}
    \centering
    \includegraphics[width=\linewidth]{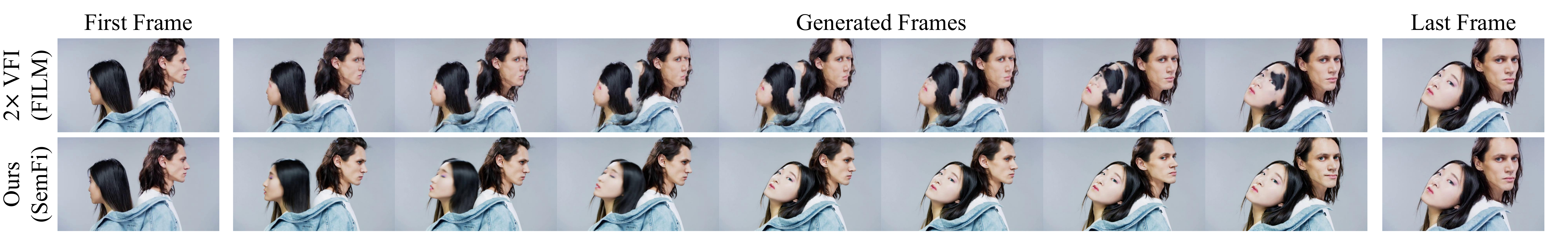}
    \caption{Qualitative comparison with conventional recursive 2$\times$ VFI. Recursive 2$\times$ VFI tends to morph the first frame into the last frame in long-sequence interpolation, while SemFi generates a more coherent intermediate evolution process consistent with the SFI objective.}
    \label{fig:11}
\end{figure*}

We also observe a clear qualitative difference between recursive 2$\times$ VFI methods and SemFi. As shown in Fig.~\ref{fig:11}, when the target sequence becomes long, especially when the two endpoint frames differ significantly, conventional 2$\times$ VFI methods tend to produce results that visually morph the first frame into the last frame. Similar phenomena can also be observed in VFI-style baselines such as Framer and FCVG. Such results mainly behave as an appearance transition between two images, rather than explicitly modeling the intermediate content evolution of behavior and actions. In contrast, SemFi better follows the objective of SFI by generating a progressive transition process from the initial state to the final state. Moreover, conventional 2$\times$ VFI methods cannot support text control, which further limits their applicability in semantic interpolation scenarios.

\subsection{Ablation Studies}

\begin{table*}[tbp]
\centering
\caption{Quantitative results in ablation studies. “w/o Multi-Frame" denotes the model trained solely on 65-frame scale data without multi-frame training; “w/o Uni. LoRA" indicates the model without the universal LoRA; "w/o Exp. LoRA" indicates the model without the expert LoRAs; "w/o Training Strategy" indicates the model obtained by training the entire MoL module jointly in a single stage.}
\begin{tabular}{lcccccccc}
\hline
        & \multicolumn{6}{c}{Video Quality} & \multicolumn{2}{c}{Semantic Consistency} \\ \cline{2-9}
Methods & Color↑ & Noise↑ & Artifact↑ & Blur↑ & Temporal↑ & Overall↑ & GLM-4.5V↑ & Qwen3-VL↑ \\ \hline
w/o Multi-Frame       & 46.83 & 47.52 & 43.48 & 43.20 & 49.73 & 45.02 & 86.20 & 94.64 \\
w/o Uni. LoRA         & 46.97 & 47.48 & 43.51 & 43.19 & 49.03 & 45.91 & 86.20 & 94.32 \\
w/o Exp. LoRA         & 47.66 & 48.37 & 44.43 & 44.10 & 50.43 & 45.96 & 86.20 & 94.34 \\
w/o Training Strategy & 48.11 & 48.91 & 44.73 & 44.41 & 51.08 & 46.33 & 86.03 & 94.14 \\ \hline
Full Model            & 49.65 & 50.37 & 46.33 & 46.00 & 52.49 & 47.95 & 86.36 & 94.39 \\ \hline
\end{tabular}
\label{tab:3}
\end{table*}

\begin{figure*}
    \centering
    \includegraphics[width=\linewidth]{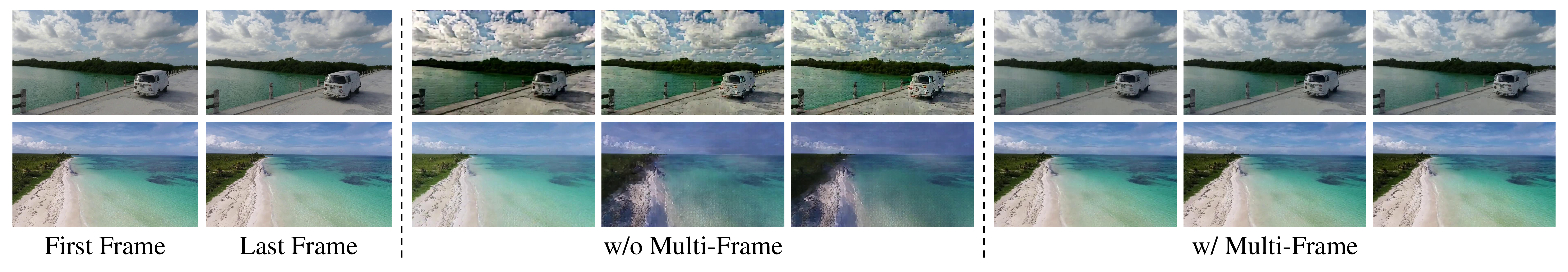}
    \caption{Ablation on training data. Generated samples for 5-frame (top) and 9-frame (bottom) cases.}
    \label{fig:5}
\end{figure*}

\begin{figure*}
    \centering
    \includegraphics[width=\linewidth]{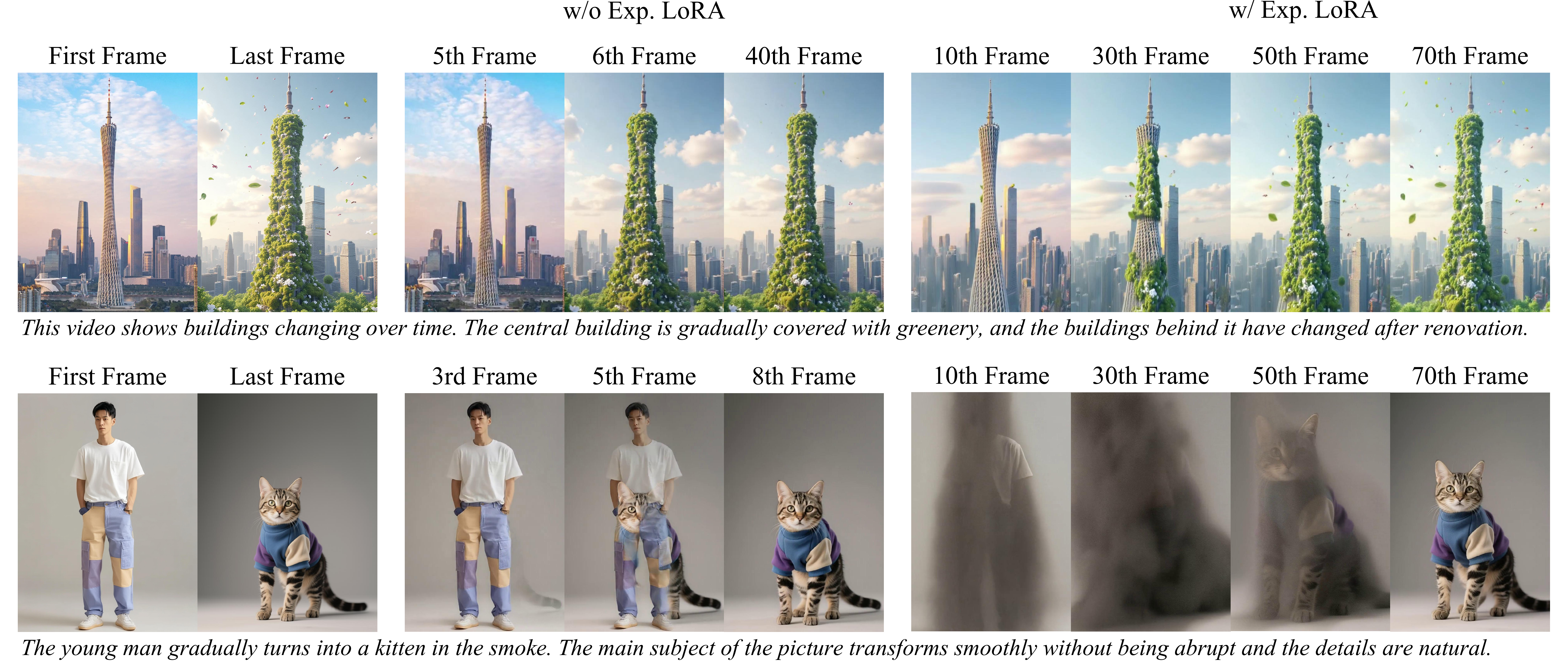}
    \caption{Ablation on expert LoRAs. Both examples show 81-frame generation results. For the "w/o Exp. LoRA" cases, we present images on frame counts with significant variations to highlight their issues.}
    \label{fig:6}
\end{figure*}

We conduct 3 groups of ablation studies to demonstrate the necessity of some key designs in our method. The quantitative results of all experiments are shown in Table~\ref{tab:3}.

\textbf{Analysis of Multi-frame Training Data.} The multi-frame training data is fundamental for achieving scale-adaptive interpolation capability. As demonstrated in Fig.~\ref{fig:5}, models trained exclusively on 65-frame videos fail to produce plausible results when handling significantly different frame counts, exhibiting artifacts like blurring and abrupt luminance changes. Single-frame training leads to overfitting to specific frame distributions, whereas multi-frame training enables the model to learn comprehensive semantic interpolation knowledge.

\textbf{Evaluation of MoL.} The MoL module, including the universal LoRA and expert LoRAs, further enhances this multi-frame interpolation capability. 1) In terms of quantitative metrics shown in Table~\ref{tab:3}, removing either type of LoRA leads to a clear drop in performance. 2) When the universal LoRA is omitted and only expert LoRAs are trained to predict interpolation within their respective ranges, the model performs poorly when the target frame count deviates significantly from the training frame count. This is due to the lack of generalized interpolation knowledge. Moreover, given the limited number of expert LoRAs, expanding the interpolation frame range inevitably causes the specialized knowledge learned for specific frame counts to fall short of covering all generation requirements. Therefore, the universal LoRA, which provides stronger generalization capability to the model, is essential. 3) Besides, Fig.~\ref{fig:6} illustrates that without expert LoRAs, the model may exhibit incorrect temporal distribution of content, resulting in abrupt transitions or unnatural acceleration. Essentially, expert LoRAs transform the frame allocation process from a closed-box operation to an explicit, controllable mechanism, preventing scale-content mismatches (e.g., generating 5-frame content across 81 frames). In summary, removing either type of LoRA directly degrades the final model’s performance, thereby verifying the necessity of the complete MoL module.

\textbf{Comparison of Different Training Strategies.} The two-stage training strategy is essential for ensuring the model reaches the expected performance. In contrast, single-stage training, where both types of LoRA are trained simultaneously, not only breaks the implicit sequential logic underlying their design but also mixes different data distributions during optimization, which reduces both training efficiency and overall effectiveness.

\subsection{Analysis of MoL Strategy}

To better understand the behavior of the proposed MoL strategy beyond standard ablations, we further conduct dedicated analyses on expert specialization and expert configuration. Since the original benchmark mainly targets final performance evaluation, the number of cases for each frame count is relatively limited. Such a setting is sufficient for overall comparison, but is less ideal for diagnosing the fine-grained behavior of experts across temporal scales. Therefore, we construct an additional analysis set specifically for the MoL study. This set is randomly sampled from unseen data sources and covers frame counts from 5 to 89 with a step size of 4, resulting in 22 different frame-count groups, and each group contains 10 cases. Compared with the original benchmark, this denser distribution provides more stable statistics for frame-count-wise analysis, allowing us to examine expert behavior under different temporal regimes in a more reliable manner. The subsequent two analyses in this subsection are evaluated based on this analysis set.

\begin{figure}
    \centering
    \includegraphics[width=\linewidth]{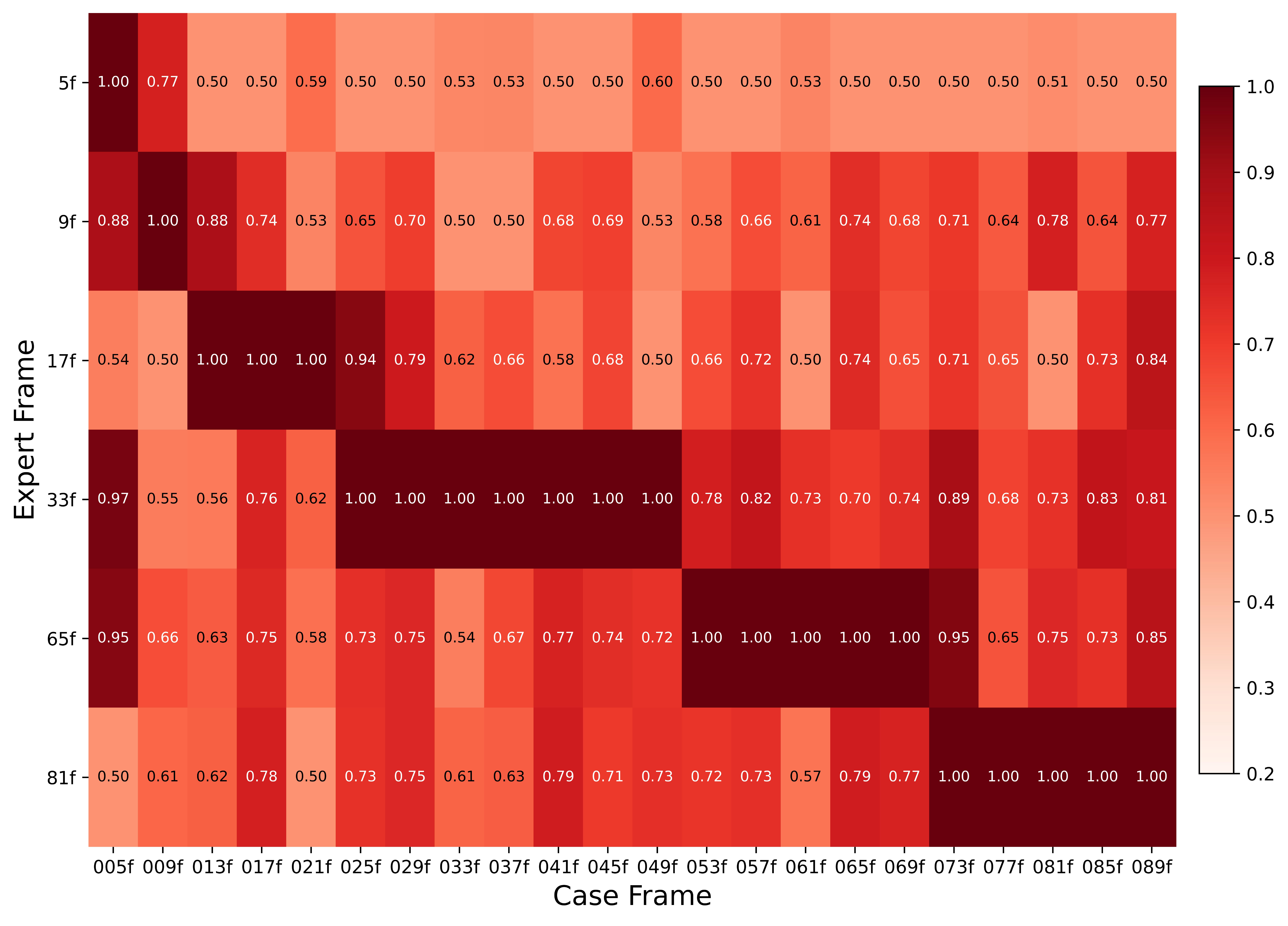}
    \caption{Cross-expert transfer heatmap of MoL. The horizontal axis denotes the target frame count of the test cases, and the vertical axis denotes the anchor frame count of the activated expert LoRA. To improve visualization while preserving relative trends, for each target frame count, we linearly normalize these 6 scores by mapping the highest value to 1.0 and the lowest value to 0.5.}
    \label{fig:9}
\end{figure}

\textbf{Cross-expert Transfer Analysis.} We first conduct a cross-expert transfer analysis to study how each expert behaves across different target frame counts. Specifically, we evaluate the analysis set multiple times, each time activating only one expert LoRA while keeping the backbone and universal LoRA unchanged. For each target frame count, we then average the "Overall" scores of all corresponding cases under each expert setting and visualize the results as a heatmap, where the vertical axis denotes the target frame count, and the horizontal axis denotes the anchor frame count of the activated expert. As shown in Fig.~\ref{fig:9}, the heatmap exhibits a clear diagonal-dominant pattern, indicating that each expert tends to perform best near its corresponding temporal scale. At the same time, the responses remain locally smooth around the diagonal, suggesting that the experts capture local temporal regimes centered around their anchor frame counts rather than isolated discrete points. This observation shows that the expert LoRAs indeed learn scale-relevant interpolation knowledge and that the proposed frame-count-aware routing is consistent with the underlying temporal structure of SFI.

\begin{figure}
    \centering
    \includegraphics[width=0.8\linewidth]{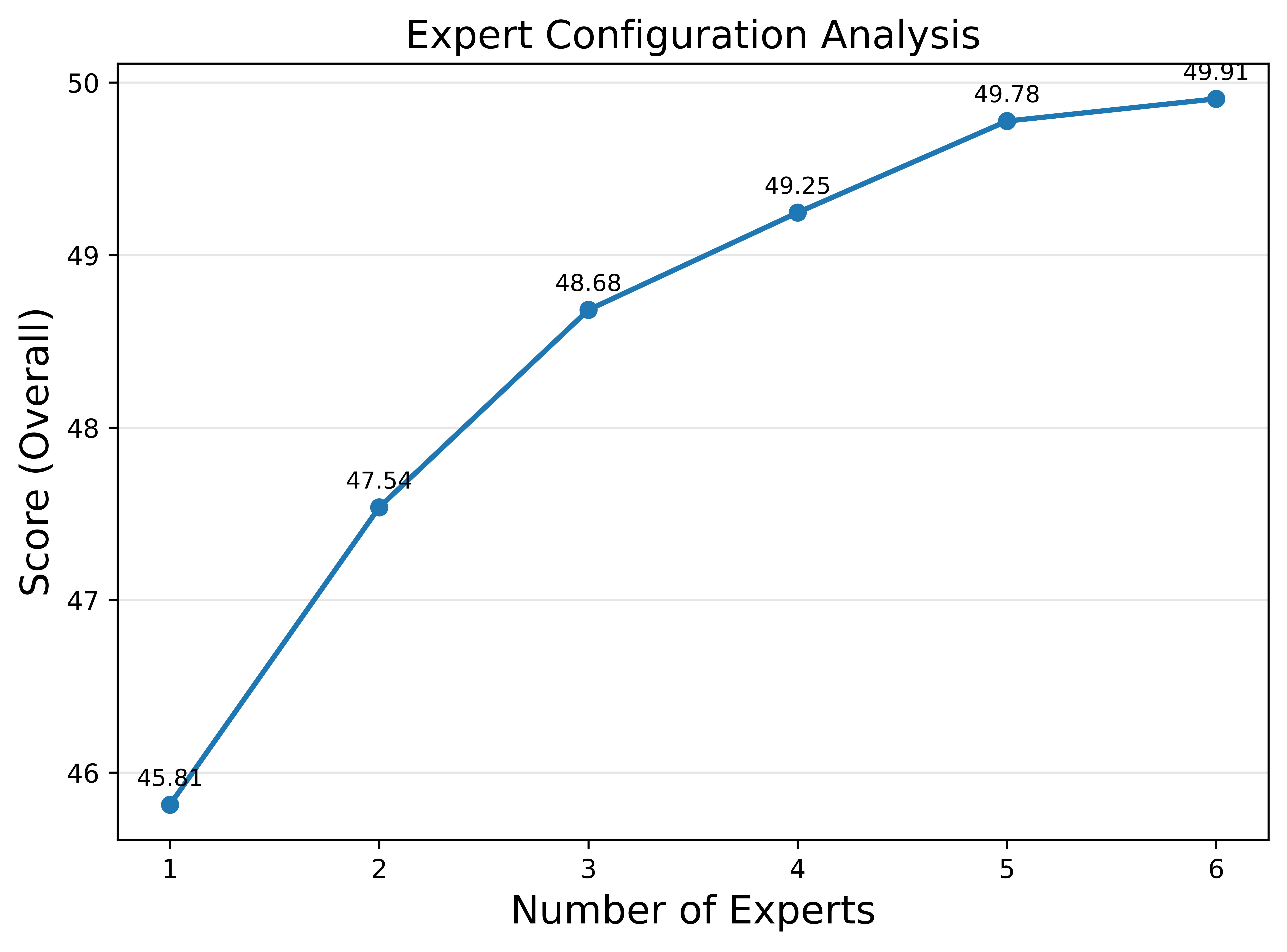}
    \caption{Overall performance under different expert configurations. The curve shows a clear monotonic improvement as the number of experts increases.}
    \label{fig:10}
\end{figure}

\textbf{Expert Configuration Analysis.} We further conduct an expert configuration analysis to study how the granularity of temporal partition affects performance. Specifically, we compare configurations with different numbers of experts while keeping the overall training strategy unchanged. The corresponding expert settings are progressively expanded from coarse to fine temporal partitioning, and the resulting "Overall" scores are reported in Table~\ref{tab:7} and Fig.~\ref{fig:10}. The results show a clear monotonic improvement as the number of experts increases, indicating that finer temporal partitioning allows the model to better capture the heterogeneity across frame scales. Combined with the cross-expert transfer analysis above, this suggests that the benefit of MoL comes from assigning experts to increasingly refined local temporal scopes: in the extreme case, the finest granularity would be to equip each frame count with its own expert. However, such a design would also introduce substantially higher training and deployment costs. Therefore, a practical expert configuration must balance specialization granularity and computational efficiency. Following the anchor-set construction principle described in Eq.~\ref{eq:1}, we finally adopt 6 experts for the current trainable range up to 81 frames, which provides a practical balance among temporal coverage, specialization ability, and computational cost.

\begin{table}[tbp]
\centering
\caption{Quantitative comparison of different expert configurations in MoL. "Expert Number" denotes the number of expert LoRAs used, "Expert Configuration" indicates the selected anchor frame counts, where \cmark\ means the corresponding MoL expert is enabled and \xmark\ means it is not used.}
\label{tab:7}
\begin{tabular}{c|cccccc|c}
\hline
\multirow{2}{*}{Expert Number} & \multicolumn{6}{c|}{Expert Configuration} & \multirow{2}{*}{Overall↑} \\ \cline{2-7}
 & \multicolumn{1}{l}{5f} & 9f & 17f & 33f & 65f & 81f &  \\ \hline
1 & \cmark & \xmark & \xmark & \xmark & \xmark & \xmark & 45.81 \\
2 & \cmark & \cmark & \xmark & \xmark & \xmark & \xmark & 47.54 \\
3 & \cmark & \cmark & \cmark & \xmark & \xmark & \xmark & 48.68 \\
4 & \cmark & \cmark & \cmark & \cmark & \xmark & \xmark & 49.25 \\
5 & \cmark & \cmark & \cmark & \cmark & \cmark & \xmark & 49.78 \\
6 & \cmark & \cmark & \cmark & \cmark & \cmark & \cmark & 49.91 \\ \hline
\end{tabular}
\end{table}

\subsection{Text-Control Capability}

\begin{figure*}
    \centering
    \includegraphics[width=\linewidth]{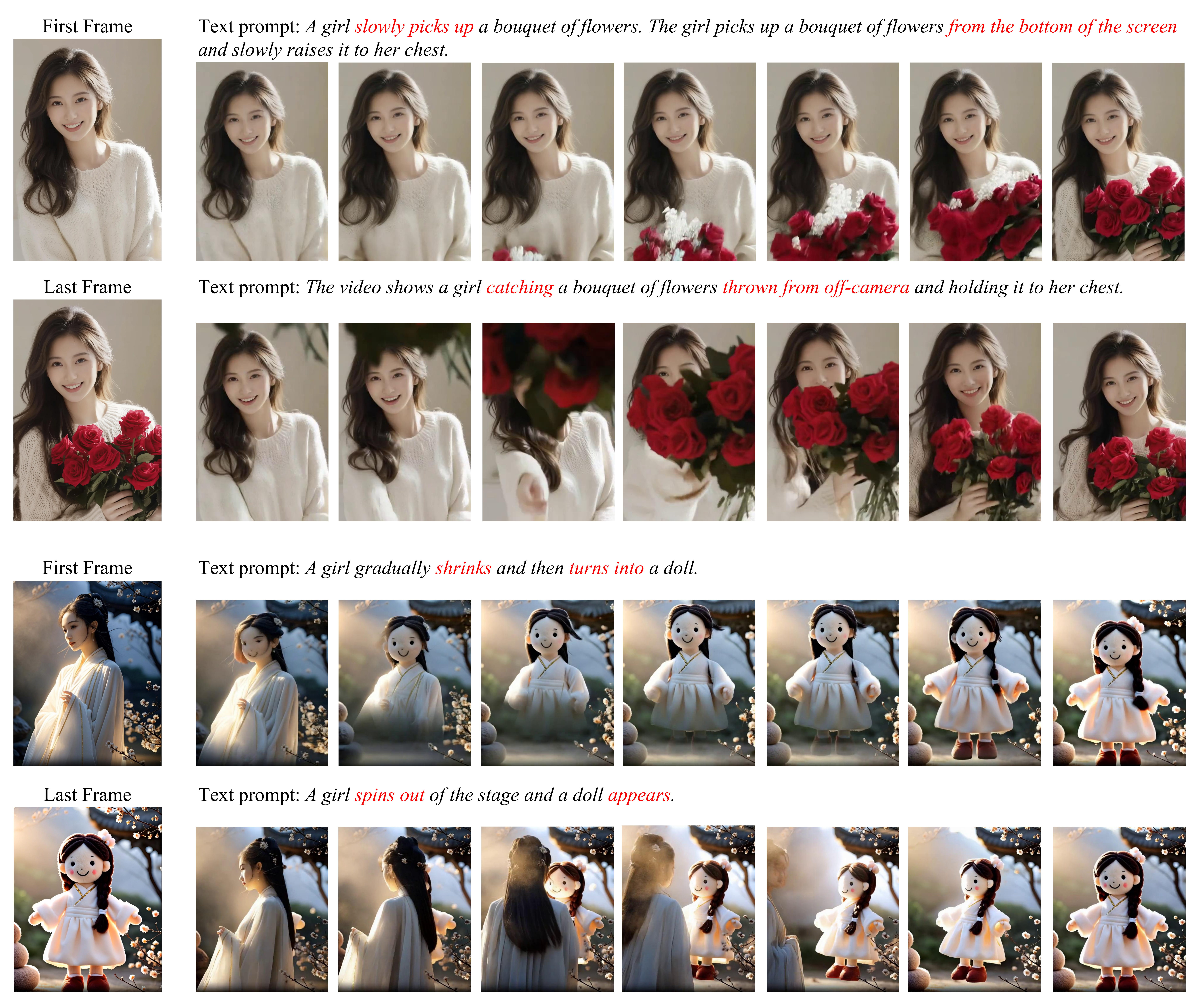}
    \caption{Results under the same start and end frames but different text prompts. SemFi demonstrates the capability to bias the content of intermediate frames according to the respective textual guidance.}
    \label{fig:7}
\end{figure*}

In the SFI task, the text prompt serves as one of the key control conditions for guiding the generation of intermediate frames, and thus should have a tangible impact on the results. Different from conventional VFI, where the first and last frames are usually temporally close and the interpolation target is largely determined by local motion continuity, SFI allows much larger temporal gaps and more diverse endpoint differences. Under such settings, multiple plausible intermediate transitions may exist between the same two endpoint frames, especially when the target frame count becomes large. Therefore, semantic control is not merely an optional extension in SFI, but a necessary mechanism for specifying the desired transition behavior and resolving ambiguity during interpolation.

To verify SemFi’s capability in semantic control, we design a validation experiment in which the first and last frames are fixed while different text prompts are provided. As shown in Fig.~\ref{fig:7}, the results demonstrate that SemFi can generate semantically coherent and visually plausible intermediate frame sequences in accordance with varying text prompts, showcasing its strong text-guided control ability. In particular, the same pair of endpoint frames can lead to clearly different intermediate evolution paths under different textual instructions, which directly reflects the necessity of semantic control in SFI.

\begin{figure*}
    \centering
    \includegraphics[width=\linewidth]{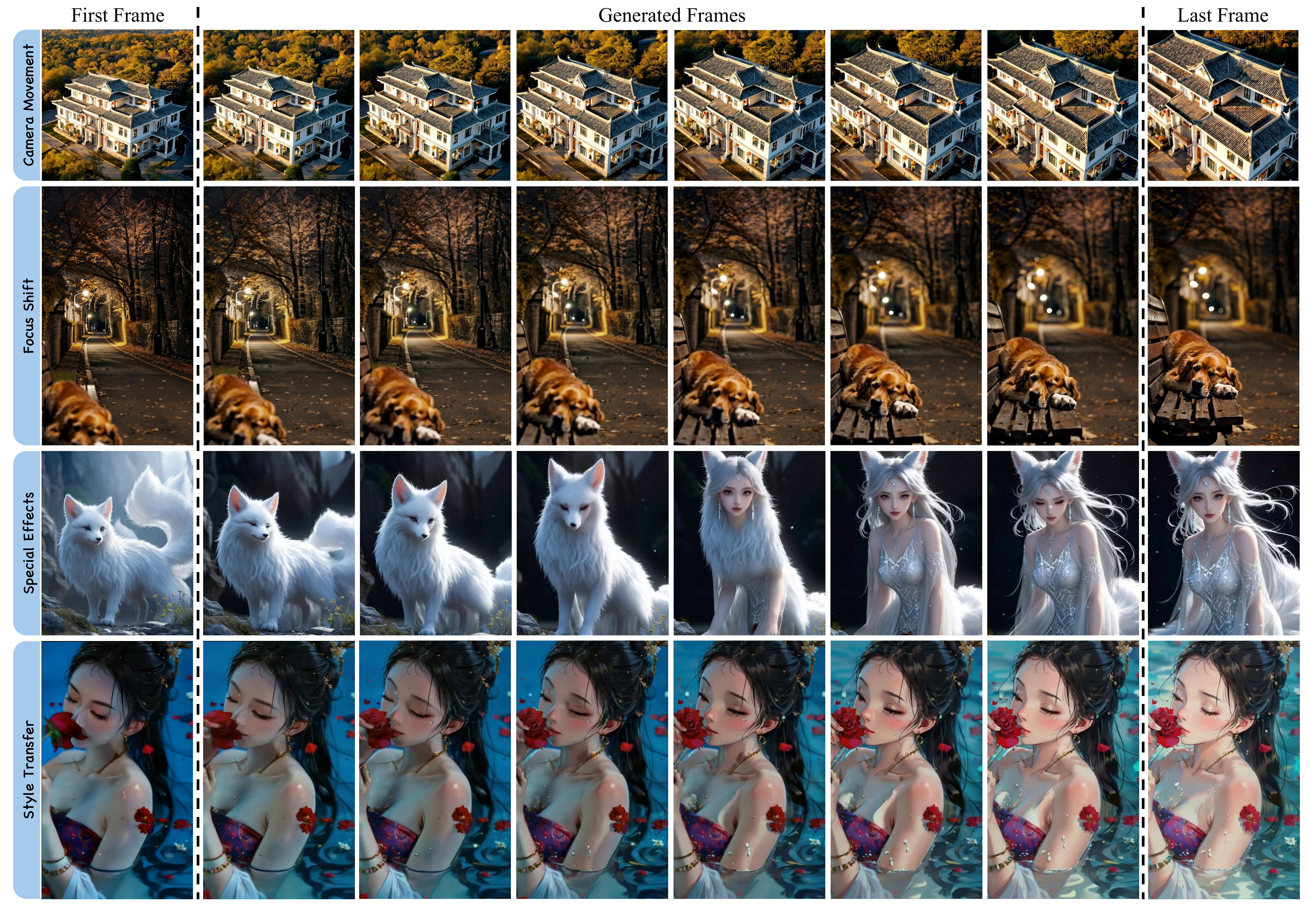}
    \caption{SemFi's interpolation results in practical scenarios. SemFi is capable of producing high-quality frames across various real-world applications.}
    \label{fig:8}
\end{figure*}

VFI methods generate content solely based on the first and last frame images, lacking any external semantic control dimension, and therefore produce fixed, non-adjustable outputs for given inputs. In contrast, SemFi introduces text prompts as an additional conditioning signal, enabling high-level semantic intervention in the interpolation process, which is particularly beneficial for maintaining content diversity when generating videos with long frame counts. This capability not only enriches the diversity of generated results but also allows the model to better accommodate complex and personalized generation requirements, offering significant advantages for creative and visually driven content generation applications.

\subsection{User Study}

To further validate the reliability of the evaluation protocol, we additionally conduct a user study on the test set. Specifically, we randomly sample 20 cases from the benchmark and collect the generated results of all 5 compared methods, resulting in 100 videos in total. We invite 16 volunteers with diverse backgrounds (including researchers in the same field with considerable expertise, researchers from other fields, and general users) to evaluate these results, and collected a total of 5,120 responses to the questions. All videos are randomly shuffled and anonymized such that participants are unaware of the corresponding generation method.

The user study consists of two parts. In the first part, for each case, participants are asked to select the most preferred result among the five methods. We then compute the average preference ratio for each method, and the results are summarized in Table~\ref{tab:5}. The results show that SemFi receives the highest overall user preference, which is consistent with the ranking given by the automatic evaluation metrics.

In the second part, participants further assign a score from 1 to 5 to each video along three dimensions: Visual Quality, which measures image clarity and the absence of noise or artifacts; Temporal Consistency, which measures whether motion is natural and free of flickering or abrupt transitions; and Prompt\&Endframe Faithfulness, which measures consistency with the text prompt and the endpoint conditions. The average scores are reported in Table~\ref{tab:6}. SemFi achieves the best overall perceptual performance across these dimensions, further confirming that the proposed method produces results that are not only favored by automatic metrics but also preferred by human observers.

Overall, the user study provides additional evidence that the automatic evaluation protocol is aligned with human perceptual judgment, while also further validating the effectiveness of SemFi from the perspective of subjective evaluation.

\subsection{Applications}

The SemFi model demonstrates extensive applicability across diverse scenarios. Primarily, it maintains full compatibility with conventional VFI tasks involving limited intermediate frames, thereby inherently supporting standard applications such as video frame rate upsampling for enhanced fluency, frame recovery, and slow-motion effect generation. Furthermore, SemFi excels in handling more challenging interpolation scenarios involving extended frame sequences and substantial semantic discrepancies between start and end reference frames. As illustrated in Fig.~\ref{fig:8}, this enhanced capacity enables the framework to address advanced use cases, including camera motion simulation, time-lapse video generation, focus transition, artistic style transfer, and special effects transformation, significantly expanding the practical utility of interpolation-based video generation systems. The capability to produce reliable and coherent interpolations under diverse practical conditions not only highlights SemFi’s strong practicality and robustness in real-world applications but also further underscores the necessity of defining the SFI task. By aligning more closely with the growing demand for diverse and controllable video content generation, SFI represents an essential step toward the next generation of intelligent video synthesis.

\subsection{Discussions and Limitations}

While SemFi has the ability to generate videos up to hundreds of frames during inference, the maximum training frame length (\textasciitilde 81) limits the upper bound of our frame set $S$, restricting application to relatively short video segments by contemporary generation standards. This constraint directly impacts both the MoL module design and the SFI-300K dataset construction. Besides, SFI-300K currently comprises predominantly real-world footage with minimal synthetic content, resulting in knowledge gaps for synthetic video interpolation scenarios. Therefore, our future research will focus on architectural enhancements to MoL for improved long-sequence adaptation and dataset refinement,  incorporating both quality-based filtering and the inclusion of carefully designed synthetic content to bridge the domain gap.

\begin{table}[tbp]
\centering
\caption{Preference study. Best results are shown in \textbf{bold}.}
\label{tab:5}
\begin{tabular}{l|c|c|c|c|c}
\hline
Method & Framer & GI & FCVG & Wan & SemFi \\ \hline
Preference & 3.75\% & 0.00\% & 8.75\% & 18.75\% & \textbf{68.75\%} \\ \hline
\end{tabular}
\end{table}

\begin{table}[tbp]
\centering
\caption{Dimension-wise perceptual rating. Best results are shown in \textbf{bold}.}
\label{tab:6}
\begin{tabular}{lccc}
\hline
 & \begin{tabular}[c]{@{}c@{}}Visual\\ Quality↑\end{tabular} & \begin{tabular}[c]{@{}c@{}}Temporal\\ Consistency↑\end{tabular} & \begin{tabular}[c]{@{}c@{}}Prompt\&Endframe\\ Faithfulness↑\end{tabular} \\ \hline
Framer & 1.93 & 2.11 & 2.31 \\
GI & 1.60 & 1.71 & 1.98 \\
FCVG & 2.68 & 2.36 & 2.68 \\
Wan & 3.90 & 3.73 & 3.91 \\ \hline
SemFi & \textbf{4.44} & \textbf{4.55} & \textbf{4.54} \\ \hline
\end{tabular}
\end{table}
\section{Conclusion} \label{sec:conclusion}

In this work, we present an advancement in video processing through the formal proposal of Semantic Frame Interpolation (SFI), a novel task paradigm that extends beyond conventional video interpolation by incorporating semantic-aware and frame-aware generation capabilities. We establish a rigorous task formulation with enhanced practical applicability, SFI-300K as the first comprehensive dataset specifically designed for SFI research, and SFIBench as a standardized evaluation framework. At last, we propose SemFi, an architecture that introduces the innovative Mixture-of-LoRA (MoL) module, enabling high-quality intermediate frame generation across diverse temporal scales.

\bibliographystyle{IEEEtran}\
\small
{
    \bibliography{main}
}

\end{document}